\documentclass[conference]{IEEEtran}
\IEEEoverridecommandlockouts
\usepackage{cite}
\usepackage{amsmath,amssymb,amsfonts}
\usepackage{algorithmic}
\usepackage{subfig} 
\usepackage{graphicx}
\usepackage{textcomp}
\usepackage{xcolor}
\usepackage{multirow}
\usepackage[hyphens]{url}
\usepackage{fancyhdr}
\usepackage{caption}
\usepackage{tablefootnote}
\def\BibTeX{{\rm B\kern-.05em{\sc i\kern-.025em b}\kern-.08em
    T\kern-.1667em\lower.7ex\hbox{E}\kern-.125emX}}

\usepackage{lipsum}
\newcommand\blfootnote[1]{%
  \begingroup
  \renewcommand\thefootnote{}\footnote{#1}%
  \addtocounter{footnote}{-1}%
  \endgroup
}

\begin{document}

\title{Lightweight Relevance Grader in RAG}

\author{\IEEEauthorblockN{Taehee Jeong}
\IEEEauthorblockA{\textit{San Jose State University} \\
taehee.jeong@sjsu.edu}

}

\maketitle

\begin{abstract}
Retrieval-Augmented Generation (RAG) addresses limitations of large language models (LLMs) by leveraging a vector database to provide more accurate and up-to-date information. When a user submits a query, RAG executes a vector search to find relevant documents, which are then used to generate a response. However, ensuring the relevance of retrieved documents with a query would be a big challenge. To address this, a secondary model, known as a relevant grader, can be served to verify its relevance. To reduce computational requirements of a relevant grader, a lightweight small language model is preferred. In this work, we finetuned llama-3.2-1b as a relevant grader and achieved a significant increase in precision from 0.1301 to 0.7750. Its precision is comparable to that of llama-3.1-70b.
Our code is available at https://github.com/taeheej/Lightweight-Relevance-Grader-in-RAG. 
\end{abstract}


\section{Introduction}
\renewcommand{\footnoterule}{%
  \kern -3pt
  \hrule width 3in height 1pt
  \kern 2pt
}
\blfootnote{2025 The 8th International Conference on Information and Computer Technologies (ICICT), IEEE Copyright 2025}

Large Language Models (LLMs) have achieved exceptional capabilities in various Natural Language Processing (NLP) tasks \cite{b1, b2, b3}, demonstrating their ability to absorb and retain vast amounts of knowledge. When responding to specific queries, LLMs often provide informative answers, leveraging the extensive range of information they acquired during their training. However, while their capabilities are impressive, LLMs still have several limitations that hinder their overall applications.

A major limitation of LLMs is the rapid growth in the number of parameters required to achieve extensive capabilities. As the training dataset expands, the model needs to capture increasingly complex patterns, which in turn demands a substantial increase in parameters. This exponential growth not only adds complexity to the model but also creates significant deployment challenges, making it difficult to implement the model in real-world applications.


Another limitation of LLMs is their inability to incorporate time-sensitive or non-public information. This limitation arises from the fact that LLMs are trained on static datasets that represent a snapshot of the internet at a particular point in time. As a result, these models often lack access to the recently developed or updated information. This can lead to a critical issue: LLMs may generate "hallucinations," where they produce responses that are not grounded in actual or current information. This problem is particularly alarming in applications where accuracy and reliability are crucial, as it can erode trust in the model's outputs.

A novel approach has recently emerged to tackle these limitations: Retrieval-Augmented Generation (RAG) \cite{b4, b5}. The RAG enhances its capabilities by integrating LLMs with external knowledge retrieval. This integration allows the RAG to access and incorporate not only publicly available information but also time-sensitive data or information that is not publicly accessible, thereby expanding its knowledge base.

\begin{figure}[ht]
\centerline{\includegraphics[width=1\linewidth]{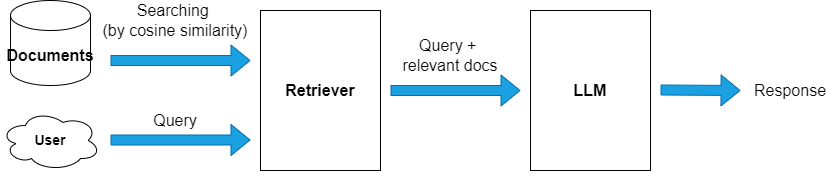}}
\caption{RAG system}
\label{fig:rag}
\end{figure}

When a query is given, RAG system uses a retriever to search an external knowledge database and retrieve the most relevant documents related to the query. Next, these documents are combined with the original query to create a prompt for a language model. The language model then generates its output based on the information from the retrieved documents, resulting in a comprehensive response to the query. The work flow of the RAG system is illustrated in Fig. \ref{fig:rag}.

\begin{figure}[ht]
\centerline{\includegraphics[width=1\linewidth]{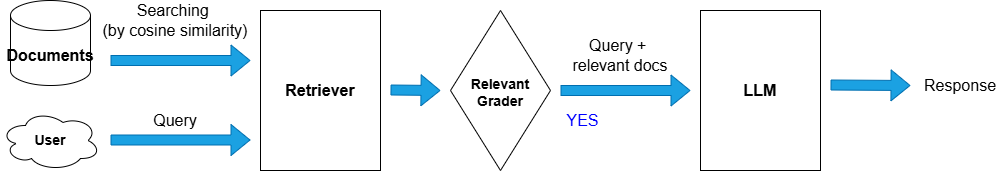}}
\caption{RAG system with a relevant grader}
\label{fig:grader}
\end{figure}

The RAG system differs from generative-only models in its ability to utilize time-sensitive information or non-public documents, such as internal company documents, to reduce the risk of hallucinations. A key component of RAG is its document retrieval mechanism, which involves comparing a query vector to document vectors in a database based on cosine similarity. The documents are ranked based on its relevance and the top matches are then selected, but this process may still yield some irrelevant documents. To refine the results, RAG employs a re-ranking process, where a secondary model acts as a relevant grader. This model assesses the retrieved documents to determine their suitability for answering the user's question, ensuring that the final response is relevant and accurate. The work flow of the RAG system with a relevance grader is illustrated in Fig. \ref{fig:grader}.

Integrating an additional LLM into the RAG pipeline poses significant memory and computational challenges. To reduce these burdens, we propose using a fine-tuned, small language model as a relevant grader. The challenge is to achieve sufficient capability with a relevant grader with a relatively small number of parameters since a language model's capability is often tied to its number of parameters [19]. Since our baseline model has only 1 billion parameters, which is significantly smaller than that of widely-used LLMs, we anticipated potential performance issues. To mitigate this, we added a binary classification head to the model's final layer, which is suitable for the binary output of a relevant grader. We then fine-tuned the model under various hyper-parameter configurations to further optimize its capabilities.

The primary contributions of this paper are as follows:
\begin{itemize}
\item Model Enhancement: A fine-tuned small language model (lama-3.2-1b) is used for relevance grading in RAG systems, improving precision from 0.1301 to 0.7756.
\item Efficiency and Speed: The lightweight model expects to minimize memory and computational requirements, enabling deployment in resource-constrained environments and accelerating the retrieval process.
\item Dataset and Evaluation: The dataset of 45,000 query-document pairs was generated for fine-tuning the relevance grading process, supporting the development of more accurate RAG systems.
\end{itemize}
Overall, this work contributes to the advancement of RAG systems by offering a practical and efficient solution for relevance grading, which enhances the accuracy and performance of information retrieval in the presence of limited computational resources.


\section{Related Work}

To identify relevant documents from a knowledge database, searching algorithms are employed in RAG systems. Traditional search algorithms rank documents by the frequency of query terms within them. Among the widely used algorithms are Term Frequency-Inverse Document Frequency (TF-IDF) and Best Matching 25 (BM25) \cite{b6}. However, these approaches primarily depend on lexical matching, which can limit their ability to effectively grasp the context of documents.


Unlike traditional search algorithms that rely on exact keyword matches, vector search utilizes vector embeddings to capture the semantics of data, enabling a meaning-based search approach. In this method, both the query and the document are independently transformed into embedding vectors using a semantic encoder. Vector search then assesses the similarity between the query vector and document vectors. This technique allows unstructured data such as images, text, and audio to be represented as vectors in a high-dimensional space, facilitating the efficient identification and retrieval of vectors that closely align with the query vector.


Distance metrics, like Euclidean distance and cosine similarity, are frequently employed to evaluate the similarity between vectors. The Euclidean distance between two embedding vectors, $\mathbf{v(s_1)}$ and $\mathbf{v(s_2)}$, each with n dimensions representing sentence 1 and sentence 2, is defined as follows:

\begin{equation}
\label{eq:euclidean}
\begin{split}
  d(s_1,s_2) &= {\left\lVert \mathbf{v(s_1)} - \mathbf{v(s_2)} \right\rVert _2} \\
  &=\sqrt{\sum_{i=0}^{n - 1}(v^1_i- v^2_i)^2}
    \end{split}
\end{equation}

Cosine similarity between two vectors $\mathbf{v(s_1)}$ and $\mathbf{v(s_2)}$  is also defined as follows.
\begin{equation}
\label{eq:cosine}
  \begin{split}
  sim(s_1,s_2) &= \frac{\mathbf{v(s_1)} \cdot \mathbf{v(s_2)}}{\left\lVert \mathbf{v(s_1)} \right\rVert _2 \left\lVert \mathbf{v(s_2)} \right\rVert _2} \\
  &= \frac{\sum_{i=0}^{n - 1} (v^1_i v^2_i)}{\sqrt{\sum{{v^1_i}^2}} \sqrt{\sum{{v^2_i}^2}}}
  \end{split}
\end{equation}

Hybrid search, a recently introduced method, integrates keyword-based search with vector-based search to capitalize on the strengths of both techniques. This combination has the potential to yield more precise and relevant search results. In a hybrid search system, keyword-based and vector-based searches are performed separately, then their results are subsequently merged. Despite its promise, one of the challenges lies in ranking these results and assigning appropriate weights to effectively combine them.

The ever-increasing volume of accessible information resources has created a significant demand for effective methods of similarity searching. Vector search algorithms are specifically developed to efficiently identify the vectors most similar to a given query vector. Among the widely used vector searching algorithms are K-Nearest Neighbors (KNN) and Approximate Nearest Neighbor (ANN).

KNN \cite{b7}, often referred to as the Brute Force algorithm, identifies the K nearest vectors to a query vector by measuring the distance—typically the Euclidean distance—between the query and every other vector in the dataset. While it ensures the precise identification of the nearest neighbors, it can be computationally demanding for large datasets.

ANN \cite{b8} algorithms permit a slight error margin, providing points that are nearly the closest rather than exactly the nearest. While this approach sacrifices some precision, it offers a substantial increase in speed over exact nearest neighbor search methods. Among the various ANN algorithms, Hierarchical Navigable Small World (HNSW) algorithm is the most widely used.

HNSW \cite{b9} creates a hierarchical graph structure, where each node corresponds to a data point and edges link nearby points in the dataset. This graph is composed of multiple layers, each representing a different level of detail or resolution. These layers are arranged hierarchically, with broader, coarser layers at the top and more detailed, finer layers at the bottom. The algorithm's main advantage lies in its ability to efficiently narrow down the search space by navigating through these layers to find the most likely candidates for nearest neighbors. This process begins at the top layer and progressively moves down to the lower layers, using the edges to steer the search towards the most similar data points. As a result, HNSW effectively balances the trade-off between search speed and accuracy.

Within RAG pipelines, re-ranking \cite{b20, b21, b22} plays a crucial role in refining initial search results to better align with user intent and context. By doing so, it enhances user satisfaction by delivering more precise and contextually relevant outcomes. This, in turn, leads to increased conversion rates and improved engagement metrics. Ultimately, re-ranking enables LLMs to leverage the most relevant and high-quality information available, resulting in more accurate and effective results.

\begin{figure}[ht]
\centerline{\includegraphics[width=1\linewidth]{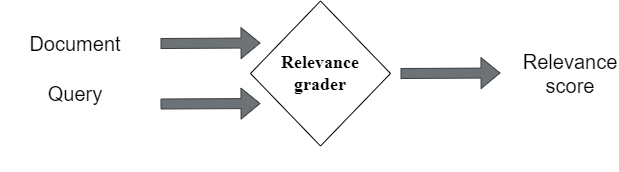}}
\caption{Re-ranking process}
\label{fig:crossencoder}
\end{figure}

Cross-encoders play a crucial role in re-ranking processes within RAG pipelines. The re-ranking process is illustrated in Fig. \ref{fig:crossencoder}. Their functionality involves taking a concatenated query and document as input, generating a relevance score as output. Although cross-encoders excel at capturing the nuanced interactions between queries and documents, their computational requirements are substantial. This is largely due to the fact that LLMs are often utilized as cross-encoders, which demands significant memory and computational resources.

The development of lightweight language models as cross-encoders seeks to strike a balance between accuracy and efficiency. With their faster processing speeds and smaller memory requirements, these models are well-suited for real-time applications. However, they often struggle to match the accuracy and contextual relevance of their larger counterparts. To address this limitation, our research focuses on developing a fine-tuned, lightweight language model that functions as a relevant grader. The goal of this model is to provide search results that are comparable in accuracy and relevance to those produced by larger, more complex language models.

\section{Dataset}
\subsection{Data Preparation}

To evaluate the accuracy of search results, we used 45,000 pairs of user queries and corresponding recent news articles. Our approach involved two main steps. First, we used a vector database of news articles which are collecting articles daily from multiple news sources \cite{b24} and embedding them using the bge-small-en-v1.5 semantic encoding model  \cite{b10}. The embedding vectors have 384-dimension.  
In the second step, we developed a set of 20 query questions across eight distinct fields: Pharmacy, Venture Capital, Information Technology (IT), Legal, Banking, Healthcare, Automotive, and Residential Construction. These queries were carefully crafted to cover various categories, including R\&D, Technology, Regulations, Market, Manufacturing, Hiring, Sustainability, Business-to-Business (B2B), Security, Industry, Leadership, Economy, and Finance. For example, one such query was \textit{"How will the expanding specialized drug market impact pharmaceutical R\&D strategy and manufacturing capabilities?"}. In total, we had 160 unique queries.

To identify relevant news articles for each query, we embedded the query sentences using the same bge-small-en-v1.5 model and calculated the cosine similarity between the query vector and the news article vectors. We utilized the HNSW algorithm for efficient vector searching, which enabled us to find the top five most similar vectors for each query. This process was repeated daily over a 90-day period for all 160 queries, resulting in the collection of 45,000 query-article pairs.

\subsection{Relevant Grading}

To guarantee that user intent and context are aligned, we combined the query with the document as input and assessed their relevance using Llama-3.1-405B-Instruct \cite{b11}. At the time of the writing of this paper, this model is the largest and most advanced openly accessible foundation model \cite{b23}. We utilized the following system prompt, incorporating chain-of-thought phrases: \textit{"Please analyze the contents of DOCUMENTS and determine whether it is relevant in answering the QUESTION"}


\begin{figure}[ht]
\centerline{\includegraphics[width=0.9\linewidth]{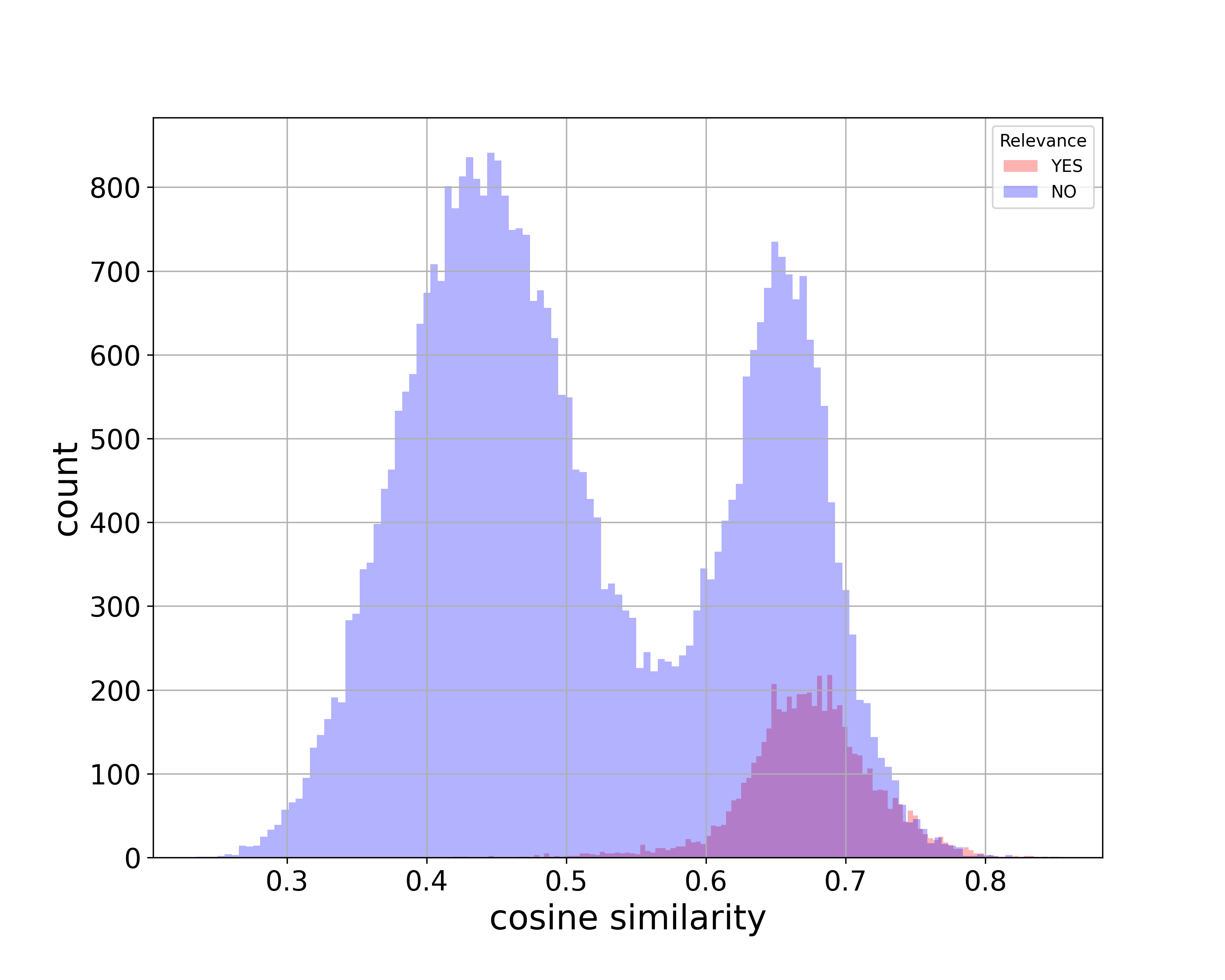}}
\caption{Distribution of cosine similarity with relevant grading}
\label{fig:cosine_sim}
\end{figure}

Fig. \ref{fig:cosine_sim} illustrates the distribution of cosine similarity along with the evaluation outcomes for relevant grading. It reveals that merely 12.3\% of the cases are approved by the relevant grader, highlighting the essential function of the relevant grader within the RAG pipeline. Additionally, the distribution displays a bimodal pattern. The second peak in this distribution corresponds to relevance, while the first peak appears to be misaligned with the relevant search. This misalignment could be attributed to the HNSW's nature as an approximate search method, which may compromise accuracy, or to an imprecise embedding model.


We evaluated its relevance outcomes against other LLMs, including GPT4o-mini \cite{b12}, Llama-3.1-70B-Instruct \cite{b13}, Llama-3.1-8B-Instruct \cite{b14}, Llama-3.2-3B-Instruct \cite{b15}, and Llama-3.2-1B-Instruct \cite{b16}. The relevance grading results from Llama-3.1-405B-Instruct were used as the Ground-True, and we calculated Accuracy, Precision, Recall, and F1-score based on the confusion matrix according to Table \ref{tab_confusion}, Eq \ref{Accuracy} - Eq \ref{F1-score}. The results are presented in Table \ref{tab_benchmark}.

\begin{table}[ht]
\caption{Confusion matrix} 
\label{tab_confusion}
\centering
\begin{tabular}{cc|cc}
 & & \multicolumn{2}{c}{\textbf{Predicted}} \\
 & & Positive & Negative \\
\cline{3-4}
\multirow{2}{*}{\rotatebox{90}{\textbf{Actual}}} 
 & Positive & TP & FN \\
 & Negative & FP & TN \\
\end{tabular}
\end{table}

\begin{equation}
\label{Accuracy}
  {Accuracy} = \frac{TP + TN}{TP+FN+FP+FN}
\end{equation}

\begin{equation}
\label{Precision}
  {Precision} = \frac{TP}{TP+FP}
\end{equation}
 
\begin{equation}
\label{Recall}
  {Recall} = \frac{TP}{TP+FN}
\end{equation}

\begin{equation}
\label{F1-score}
  {F_1} = \frac{2}{\frac{1}{Precision}+\frac{1}{Recall}}
\end{equation}

\begin{table}[ht]
\caption{Model comparison for relevance grading} 
\label{tab_benchmark}
\begin{center} 
\scalebox{0.95}
{
\begin{tabular}{|c|c|c|c|c|}
\hline
\rule[-1ex]{0pt}{3.5ex}  Model  & Accuracy & Precision & Recall & F1-score\\
\hline\hline
\rule[-1ex]{0pt}{3.5ex}  gpt4o-mini & 0.9302 & 0.7170 & 0.7118 & 0.7144  \\
\hline
\rule[-1ex]{0pt}{3.5ex}  llama3.1-70b & 0.9582 & 0.8341 & 0.8225 & 0.8283 \\
\hline
\rule[-1ex]{0pt}{3.5ex}  llama3.1-8b & 0.8815 & 0.5116 & 0.7607 & 0.6118 \\
\hline
\rule[-1ex]{0pt}{3.5ex}  llama3.2-3b & 0.7793 & 0.3385 & 0.8372 & 0.4820 \\
\hline
\rule[-1ex]{0pt}{3.5ex}  llama3.2-1b & 0.2367 & 0.1312 & 0.9288 & 0.2299 \\
\hline
\end{tabular}\vspace{-20pt}
}
\end{center}
\end{table}


The dataset is imbalanced, comprising a majority of negative labeled data. This imbalance can result in a high false positive rate in a model's predictions. To effectively evaluate the model's performance in this context, Precision is a particularly useful metric, as it helps assess the accuracy of the model's positive predictions. As anticipated, a model with a large number of parameters, like Llama-3.1-70B, achieves the highest Precision score of 0.8341. In contrast, a model with fewer parameters, such as Llama-3.2-1B, has the lowest Precision score of 0.1312, in line with scale's law \cite{b19}. Despite Llama-3.2-1B having the poorest Precision score among the other models, it is suitable for efficient deployment in RAG systems due to its lightweight design, which requires less memory and computing operations. Our objective in this work is to fine-tune Llama-3.2-1B to enhance its Precision, enabling it to function effectively as a relevant grader.

\section{Task-specific fine-tuning}
\label{Experiment}

Fine-tuning a language model on specialized data allows it to leverage its extensive pre-learned knowledge and adapt to a specific task. By modifying its parameters through fine-tuning, the model can better align with the demands of the task, resulting in improved performance and applicability within that domain. This approach is particularly effective when we want to optimize the model's performance for a single, well-defined task, ensuring that the model excels in generating task-specific content with precision and accuracy.

\begin{figure}[ht]
\begin{center}
  \subfloat[][Full fine-tuning]{\includegraphics[width=0.8\linewidth]{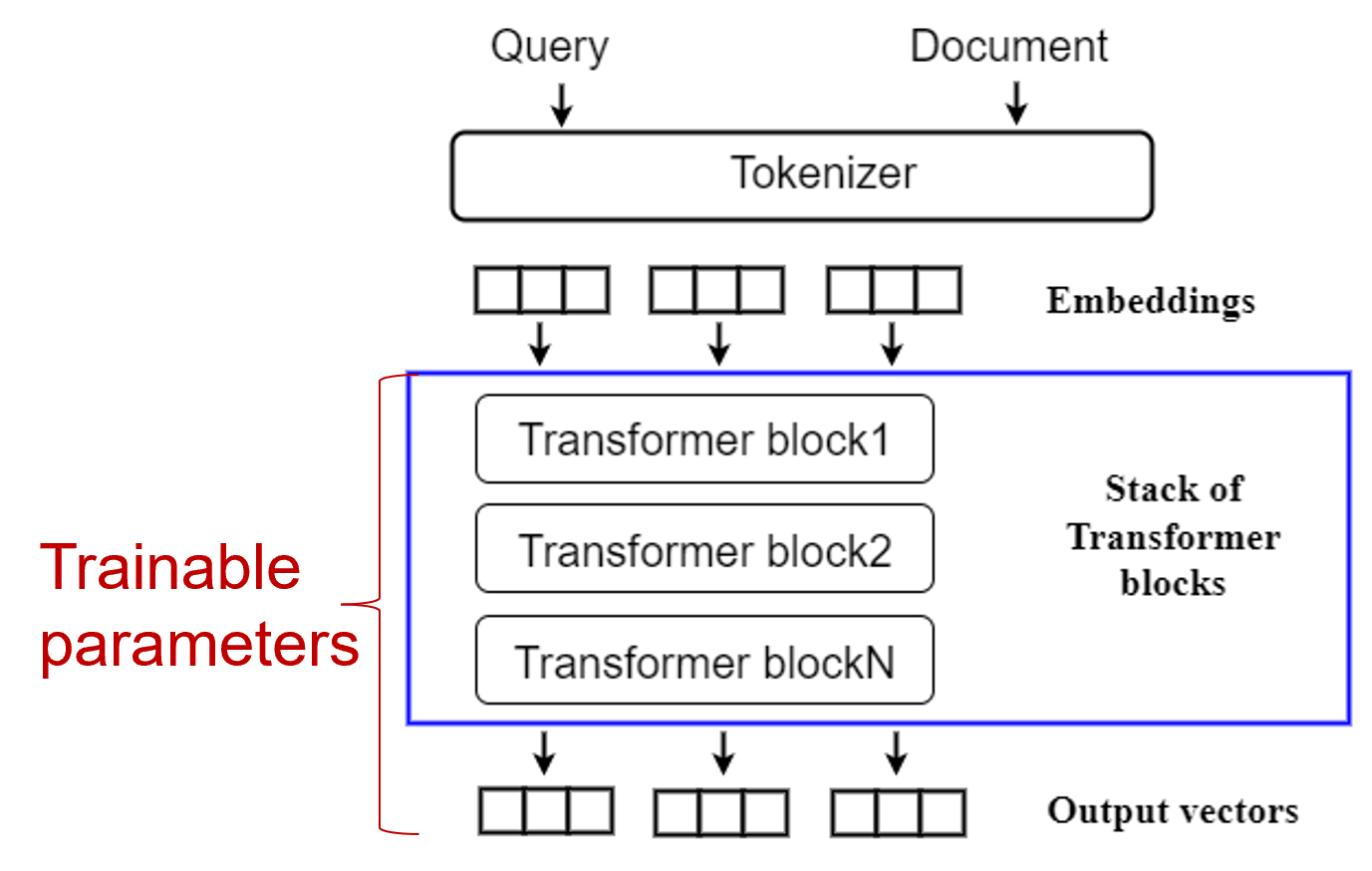}}
    \vfil
  \subfloat[][Transfer learning with Classification head]{\includegraphics[width=0.8\linewidth]{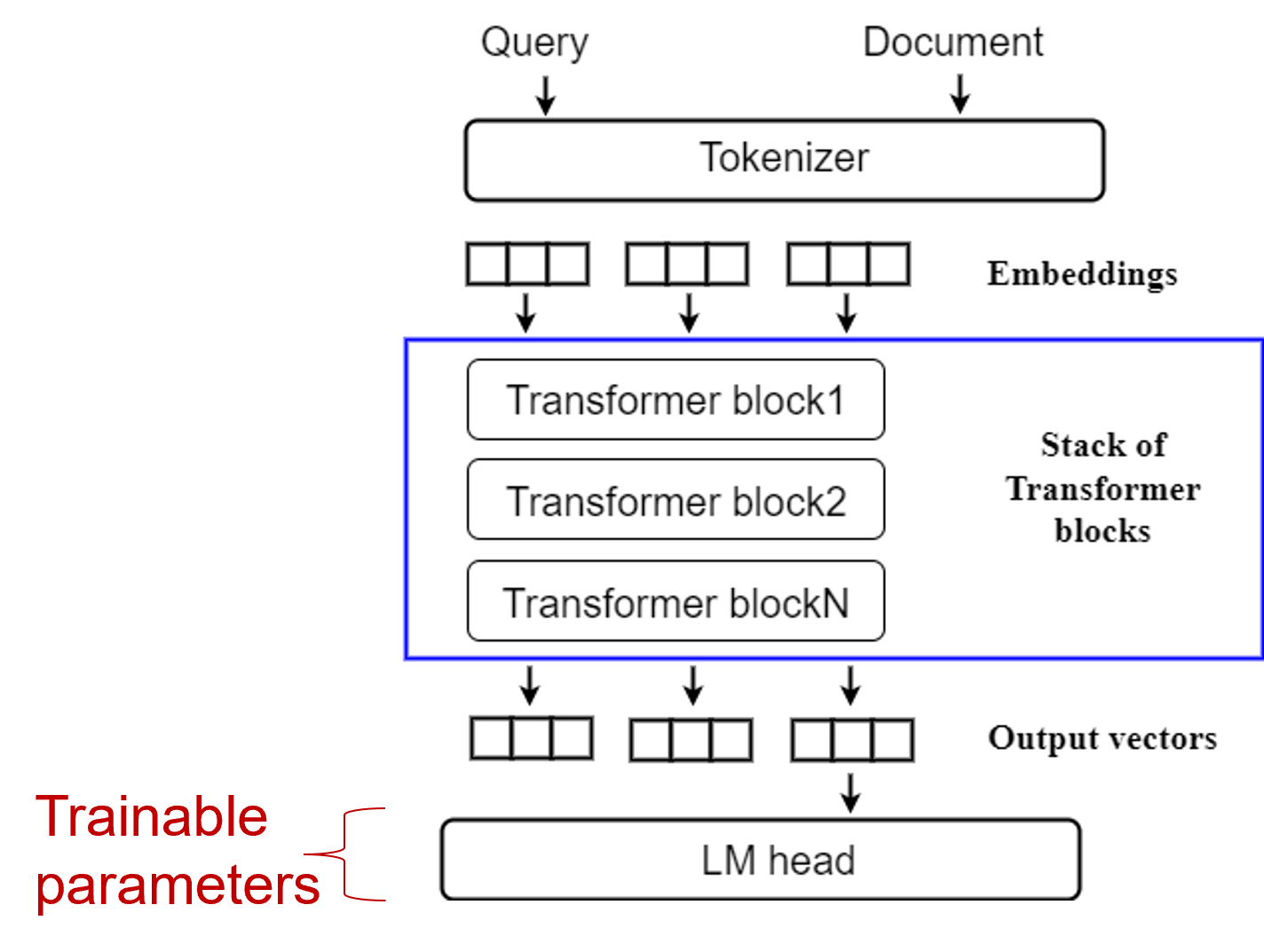}}
  
  \subfloat[][Full fine-tuning with Classification head]{\includegraphics[width=0.8\linewidth]{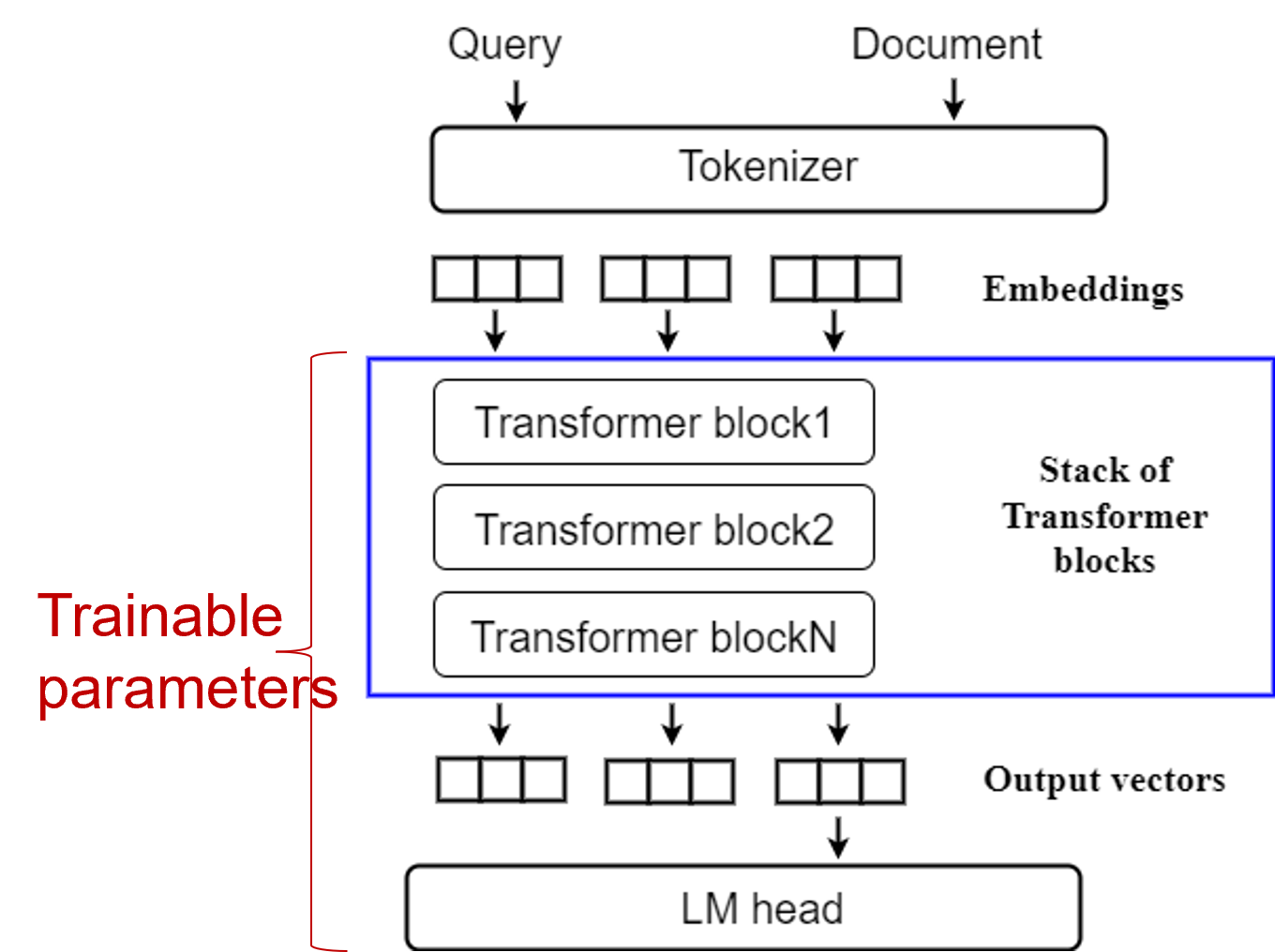}}

\end{center}
\caption{Model Configuration for Fine-tuning} 
\label{fig:exp}
\end{figure}

Our work began with the Llama-3.2-1B model as our foundation. We aimed to fine-tune this baseline model to perform as a relevant grader, a task that requires assessing the relevance between a user's query and a set of documents. Specifically, the model would take a user's query and a related document as input and output a determination of whether the query and the document are relevant. To avoid overfitting, we divided the dataset of 45,000 user query and document pairs into 80\% for training and 20\% for testing. The training and testing datasets preserve the same proportion of positive to negative labels. 





The fine-tuning process involves adjusting the parameters of a model to better suit a specific task. The degree of modification can vary greatly depending on the task's requirements. Model configurations for fine-tuning are illustrated in Fig.\ref{fig:exp}.

\subsection{Full fine-tuning}
\label{fft}


Full fine-tuning involves a comprehensive adjustment of a model, where all parameters of its layers are modified using data that is specifically tailored to a particular task. 
In our case, we fine-tuned every layer of Llama-3.2-1B-Instruct using a training dataset consisting of 36,000 pairs of user query and document.

\begin{table}[ht]
\caption{Model comparison on test dataset} 
\label{tab_finetune1}
\begin{center} 
\scalebox{0.9}
{
\begin{tabular}{|c|c|c|c|c|}
\hline
\rule[-1ex]{0pt}{3.5ex}  Model  & Accuracy & Precision & Recall & F1-score\\
\hline\hline
\rule[-1ex]{0pt}{3.5ex}  llama3.1-70b & 0.9573 & 0.8239 & 0.8299 & 0.8269 \\
\hline
\rule[-1ex]{0pt}{3.5ex}  llama3.1-8b & 0.8848 & 0.5208 & 0.7710 & 0.6217 \\
\hline
\rule[-1ex]{0pt}{3.5ex}  llama3.2-3b & 0.7722 & 0.3309 & 0.8371 & 0.4744 \\
\hline
\rule[-1ex]{0pt}{3.5ex}  llama3.2-1b & 0.2367 & 0.1301 & 0.9176 & 0.2279 \\
\hline
\rule[-1ex]{0pt}{3.5ex}  Config.A & 0.7439 & 0.1655 & 0.2688 & 0.2049 \\
\hline
\rule[-1ex]{0pt}{3.5ex}  Config.B  & 0.5730 & 0.1411 & 0.4869 & 0.2187 \\
\hline
\rule[-1ex]{0pt}{3.5ex}  Config.C & 0.9353 & 0.7750 & 0.6670 & 0.7170 \\
\hline
\end{tabular}\vspace{-20pt}
}
\end{center}
\end{table}

We utilized the AdamW optimizer \cite{b17} with a cosine learning rate schedule. The schedule started with an initial learning rate of 2e-5 and gradually decreased to a final learning rate that was 10\% of the peak rate. Cross-entropy was used as the loss function. Since the training dataset was skewed, with a predominance of negative labels, we implemented both oversampling and under-sampling techniques to achieve a more balanced distribution of positive and negative labels, thereby mitigating the impact of class imbalance on our model's performance.

\subsection{Transfer learning with Classification head}
\label{transfer}

Our study employed transfer learning, a technique that harnesses knowledge gained from one task or pre-existing knowledge obtained through pre-training on a large dataset to enhance performance on a specific task. To implement this approach, we leveraged a pre-trained Llama model and attached a classification head, a specialized layer designed for classification tasks, to its end. The classification head plays a crucial role in predicting the final label by processing the model's output. Specifically, it takes the hidden state with a dimension of 2048 and converts it into a logit with a dimension of 2, corresponding to the number of labels. The logit then undergoes softmax and argmax processing to yield the final label. A significant benefit of this transfer learning approach is the substantial reduction in computational operations required during training. By utilizing a pre-trained model, we avoided the need to train a large model with 1.236 billion parameters, instead training only a single classification layer with 4096 parameters, resulting in considerable computational savings.

\subsection{Full fine-tuning with Classification head}



In the previous section \ref{transfer}, we explored a method where a pre-trained LLM was used as a fixed feature extractor, with a classification head appended to its end for a specific classification task. The pre-trained LLM provided comprehensive representations, which were then tailored to the task at hand by training the final layers on a relevance grading dataset, while keeping the rest of the model unchanged. This approach allowed for efficient fine-tuning of the LLM. However, despite observing an improvement in precision, the results did not fully meet our expectations. To further improve performance, we also experimented with fully fine-tuning the model, including the addition of a classification head, which involved training the entire model parameters on task-specific data. Unlike the previous approach, which only modified the final layers, full fine-tuning adjusted all model layers during training.



After fine-tuning the model, we evaluated its performance on the test dataset by measuring Accuracy, Precision, Recall, and F1-score, and compared these metrics with other language models, as shown in Table \ref{tab_finetune1}. Fully fine-tuned Llama-3.2-1B (Configuration A) demonstrated an improvement in Precision, increasing from 0.1331 to 0.1655, although it still lags behind the Precision of Llama-3.2-70B. Fully fine-tuned llama-3.2-1b with a classification head (Configuration C) achieved a Precision of 0.7750, which is significantly higher than that of llama3.1-8b and GPT4o-mini, but slightly below that of llama3.1-70b.

\begin{figure}[ht]
\centerline{\includegraphics[width=0.9\linewidth]{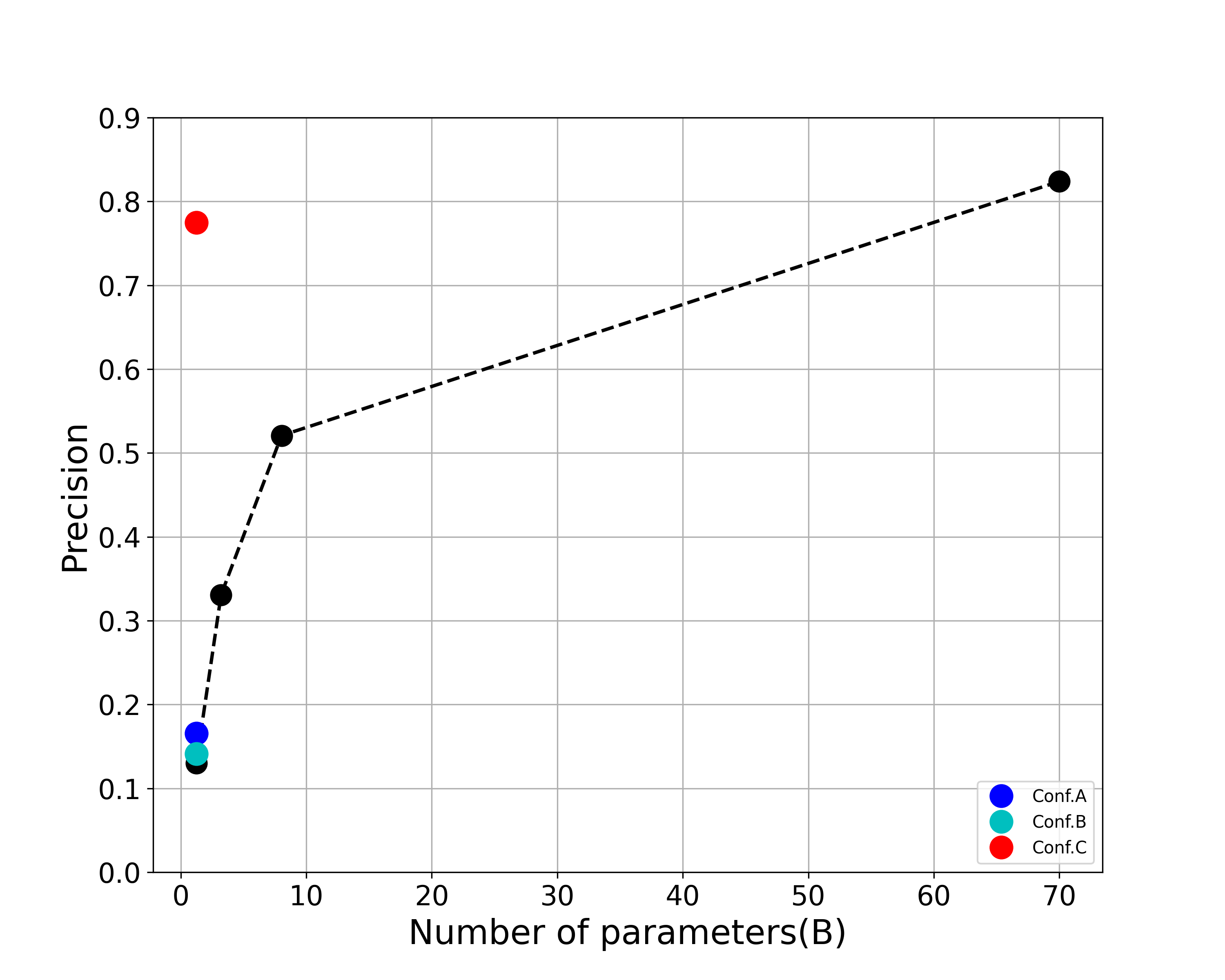}}
\caption{Precision of relevance grading on test dataset}
\label{fig:Precision}
\end{figure}

The relationship between model complexity and precision is illustrated in Fig. \ref{fig:Precision}, which shows that models with a larger number of parameters generally tend to achieve higher precision on the test dataset. Our full fine-tuned llama-3.2-1b model with a classification head, demonstrated particularly impressive results. Notably, it exceeded the typical performance expectations outlined by the scale's law, suggesting that our approach can lead to exceptional outcomes.



\section{Conclusion}

Recently, RAG has emerged as a novel approach to address the limitations of LLMs. One major issue with LLMs is their tendency to provide outdated information and generate hallucinations, leading to incorrect or misleading responses. To resolve this, RAG leverages similarity search to retrieve relevant data from a vector database that encompasses a wide range of media, including images, audio, video, and text. The core of RAG relies on cosine similarity to search for related documents, which is crucial for information retrieval. However, the documents retrieved through cosine similarity sometimes fail to fully address the query, necessitating a re-ranking process to assess their relevance. 

To refine the search results, a re-ranking process is employed, which typically involves an additional language model as a relevance grader. However, this approach is resource-intensive, requiring significant memory and computational operations within RAG pipelines. A potential solution is to utilize a lightweight language model as a relevant grader, but a lightweight model often under-perform. To overcome this challenge, we fine-tuned small language model for the relevance grading task, which aims to strike a balance between efficiency and accuracy.

To develop an effective relevant grader for a RAG system, we generated a dataset of 45,000 query-document pairs. We started with the lama-3.2-1b model as our baseline and enhanced it by adding a classification head layer. Through fine-tuning with classification head, we significantly improved the model's Precision from 0.1301 to 0.7756, making it suitable for relevance grading.

The use of a lightweight language model like lama-3.2-1b offers several advantages in a RAG system. Firstly, it reduces the demand for memory and computational operation, making it more feasible to implement a RAG system in resource-constrained environments. Additionally, the smaller model size enables faster computation, accelerating the retrieval process. Overall, our approach enables the efficient and effective implementation of re-ranking process in a RAG system, striking a balance between performance and resource requirements.

\section*{Appendix A. Study of Loss function}

There are two distinct types of loss functions used in classification: cross-entropy loss and contrastive loss \cite{b18}. Cross-entropy loss evaluates the class probabilities generated by a model for each sample individually, without imposing any constraints on the model's internal representation of the input. In contrast, contrastive loss operates on pairs of embeddings produced by the model, along with a Boolean label indicating whether the two vectors are similar or not.

The key difference between these two loss functions lies in how they handle similar and dissimilar pairs during training. Contrastive loss penalizes similar pairs if they are too far apart, based on Euclidean distance, while dissimilar pairs are penalized if they are too close. To achieve this, contrastive loss introduces a concept called a "margin," which is the minimum distance that dissimilar points must maintain. If the distance between two vectors is less than the margin, it suggests that the pair is likely relevant, whereas a distance greater than the margin indicates that the vectors probably represent different pairs.

We employed both Contrastive loss and Cross-entropy as our loss functions. The Contrastive loss is defined as following. 

\begin{equation}
  {Loss} = (1-Y)\frac{1}{2}(D_w)^2 + Y\frac{1}{2}[max(0,m-D_w)]^2
\end{equation}
where $D_w$ is Euclidean distance between the two input vectors, m $>$ 0 is a margin, and Y is a binary label assigned to this pair.

\begin{table}[ht]
\caption{comparison by Loss function on test dataset} 
\label{tab_transfer}
\begin{center} 
\scalebox{0.95}
{
\begin{tabular}{|c|c|c|c|c|c|}
\hline
\rule[-1ex]{0pt}{3.5ex}  Config  & Loss func & Accuracy & Precision & Recall & F1-score\\
\hline\hline
\rule[-1ex]{0pt}{3.5ex}  B & Cross-entropy & 0.5730 & 0.1411 & 0.4869 & 0.2187 \\
\hline
\rule[-1ex]{0pt}{3.5ex}  B & Contrastive & 0.7356 & 0.1657 & 0.2860 & 0.2098 \\
\hline
\rule[-1ex]{0pt}{3.5ex}  C & Cross-entropy & 0.9353 & 0.7750 & 0.6670 & 0.7170 \\
\hline
\rule[-1ex]{0pt}{3.5ex}  C & Contrastive & 0.9291 & 0.7256 & 0.6796 & 0.7018 \\
\hline
\end{tabular}\vspace{-20pt}
}
\end{center}
\end{table}

In this study, we represented both the model's output and the actual label as vectors. We then calculated the Euclidean distance between these two vectors, using a value of 1 for the parameter m. Table \ref{tab_transfer} presents a comparison of the fine-tuned model's performance using Cross-entropy loss versus Contrastive loss. The model's Precision with Contrastive loss surpasses that with Cross-entropy loss in the case of configuration of B, but lags behind in the case of configuration of C.

\section*{Appendix B. Parameter-Efficient Fine-Tuning}

\begin{table}[ht]
\caption{comparison by rank and alpha on test dataset} 
\label{tab_lora}
\begin{center} 
\scalebox{0.95}
{
\begin{tabular}{|c|c|c|c|c|c|}
\hline
\rule[-1ex]{0pt}{3.5ex}  rank  & alpha & Accuracy & Precision & Recall & F1-score\\
\hline\hline
\rule[-1ex]{0pt}{3.5ex}  16 & 16 & 0.8766 & 0.2500 & 0.0027 & 0.0054 \\
\hline
\rule[-1ex]{0pt}{3.5ex}  16 & 32 & 0.8763 & 0.1000 & 0.0009 & 0.0018 \\
\hline
\rule[-1ex]{0pt}{3.5ex}  16 & 64 & 0.8747 & 0.2051 & 0.0072 & 0.0140 \\
\hline
\rule[-1ex]{0pt}{3.5ex}  16 & 128 & 0.8771 & 0.3333 & 0.0009 & 0.0018 \\
\hline
\rule[-1ex]{0pt}{3.5ex}  128 & 16 & 0.8749 & 0.2162 & 0.0072 & 0.0140 \\
\hline
\rule[-1ex]{0pt}{3.5ex}  128 & 32 & 0.8766 & 0.2857 & 0.0036 & 0.0071 \\
\hline
\rule[-1ex]{0pt}{3.5ex}  128 & 64 & 0.8757 & 0.1500 & 0.0027 & 0.0053 \\
\hline
\rule[-1ex]{0pt}{3.5ex}  128 & 128 & 0.8716 & 0.1772 & 0.0127 & 0.0236 \\
\hline
\rule[-1ex]{0pt}{3.5ex}  256 & 256 & 0.8594 & 0.1226 & 0.0235 & 0.0395 \\
\hline
\end{tabular}\vspace{-20pt}
}
\end{center}
\end{table}

Parameter Efficient Fine-Tuning (PEFT) reduces the number of fine-tuning parameters and memory usage during fine-tuning process of a pre-trained large model \cite{b25, b26}. Among various PEFT algorithms, Low-Rank Adaptation (LoRA) is widely used, which freezes the pretrained weights of LLMs and trains newly generated rank decomposition matrices \cite{b27}.
We utilized LoRA to finetune our baseline model with Supervised Fine-Tuning Trainer (SFTTrainer) \cite{b28}. The initial learning rate is 5e-5 and a cosine
learning rate schedule is applied. Table \ref{tab_lora} presents a comparison of the fine-tuned model's performance with a various combination of lora rank and alpha, a scaling parameter. Overall, the Recall value of LoRA is very low, which indicates poor performance compared to that of full fine-tuning.

\end{document}